%% file: main.tex
\documentclass[10pt,twocolumn,letterpaper]{article}

\usepackage{cvpr}
\usepackage{times}
\usepackage{epsfig}
\usepackage{graphicx}
\usepackage{amsmath}
\usepackage{amssymb}

\usepackage{url}
\usepackage{xcolor}
\usepackage{epsfig} 

\input{text/macros}

\usepackage[pagebackref=true,breaklinks=true,letterpaper=true,colorlinks,bookmarks=false]{hyperref}

\cvprfinalcopy 

\def\cvprPaperID{4} 

\makeatletter
\def\blfootnote{\xdef\@thefnmark{}\@footnotetext}
\makeatother

\ifcvprfinal\fi
\begin{document}

\title{\LARGE \bf Live Reconstruction of Large-Scale Dynamic Outdoor Worlds}

\author{Ondrej Miksik$^*$\\
Emotech Labs
\and
Vibhav Vineet$^*$\\
Microsoft Research
}

\maketitle

\begin{abstract}
Standard 3D reconstruction pipelines assume stationary world, therefore suffer from ``ghost artifacts'' whenever dynamic objects are present in the scene. 
Recent approaches has started tackling this issue, however, they typically either only discard dynamic information, represent it using bounding boxes or per-frame depth or rely on approaches that are inherently slow and not suitable to online settings.

We propose an end-to-end system for live reconstruction of large-scale outdoor dynamic environments.\blfootnote{$^*$ Equal contribution.}
We leverage recent advances in computationally efficient data-driven approaches for 6-DoF object pose estimation to segment the scene into objects and stationary ``background''. 
This allows us to represent the scene using a time-dependent (dynamic) map, in which each object is explicitly represented as a separate instance and reconstructed in its own volume.
For each time step, our dynamic map maintains a relative pose of each volume with respect to the stationary background.
Our system operates in incremental manner which is essential for on-line reconstruction, handles large-scale environments with objects at large distances and runs in (near) real-time.
We demonstrate the efficacy of our approach on the KITTI dataset, and provide qualitative and quantitative results showing high-quality dense 3D reconstructions of a number of dynamic scenes.
\end{abstract}


\input{text/introduction}
\input{text/related_work}

\input{text/method}

\input{text/results}

\input{text/conclusion}

{\small
\bibliographystyle{ieee}
\bibliography{main}
}

\end{document}

%% file: text/macros.tex
\newcommand\ignore[1]{}

\newcommand{\figref}[1]{Fig.~\ref{#1}}

\def \path{\bp C}

\newcommand{\bft}{{\mathbf{t}}}

\newcommand{\bfP}{{\mathbf{P}}}

\newcommand{\bfR}{{\mathbf{R}}}

\newcommand{\bfV}{{\mathbf{V}}}

\newcommand{\calP}{{\mathcal{P}}}

\newcommand{\calV}{{\mathcal{V}}}
\newcommand{\calW}{{\mathcal{W}}}

\newcommand{\bbR}{{\mathbb{R}}}

%% file: text/introduction.tex
\vspace{-0.25cm}
\section{Introduction}

The world around us represents an inherently \emph{dynamic} 3D environment.
Hence all robots operating in this complex environment need not just to recognise the stationary parts of it, but also continuously perceive and reason about all other dynamically moving agents or objects. 
For instance, autonomous cars need to understand geometric and spatial extent of all other moving cars, pedestrians or bicyclists, and reason about their actions and interactions in order to move safely around them.
At the core of this understanding lies accurate 3D reconstruction of each object (and stationary background) -- a fundamental computer vision problem called ``multi-body dynamic scene reconstruction''.

\begin{figure}[t]
    \centering
    \includegraphics[width=\linewidth]{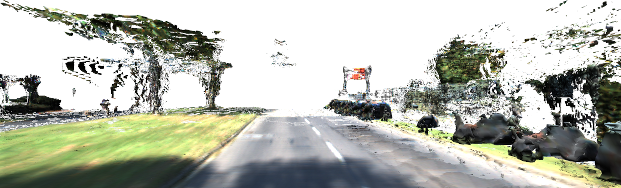} \vspace{0.2cm}
    \includegraphics[width=\linewidth]{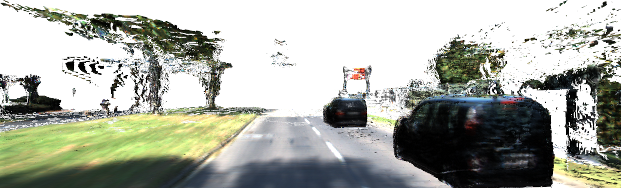} \vspace{0.2cm}
    \includegraphics[width=\linewidth]{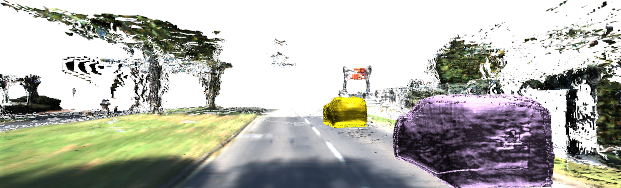}
    \vspace{-0.55cm}
    \caption{Static 3D reconstruction suffers from ``ghost artifacts'' caused by dynamic objects (top). Our system (middle) explicitly models each object instance as a separate volume (bottom). Live output of our system, as seen from a moving platform on-the-fly.} 
    \label{fig:teaser}
    \vspace{-0.6cm}
\end{figure}

Most early approaches to (dense) Structure-from-Motion (SfM) or Visual SLAM focused on 3D reconstruction of static parts of the scene~\cite{rome_in_a_day, NewcombeIHMKDKSHF_ismar11}. 
In other words, the dynamic component of the scene was considered to be noise which had to be (explicitly) suppressed to prevent the well-known \emph{ghost artifact} that generally appears whenever dynamic content is fused into the stationary volume~\cite{dynslam, VineetMLNGPKMIP_icra15}. 
This approach is useful for reconstruction of (largely) stationary worlds such as museums or galleries~\cite{reconstructing_museums, qian_yi2013iccv} where the goal is to produce a dense \emph{static} 3D model which is not corrupted by moving objects.
However, discarding dynamic information is absolutely unacceptable for decision making of \emph{any} agent operating in dynamic environments.

\begin{figure*}[t]
    \centering
    \includegraphics[width=\linewidth, height=5cm]{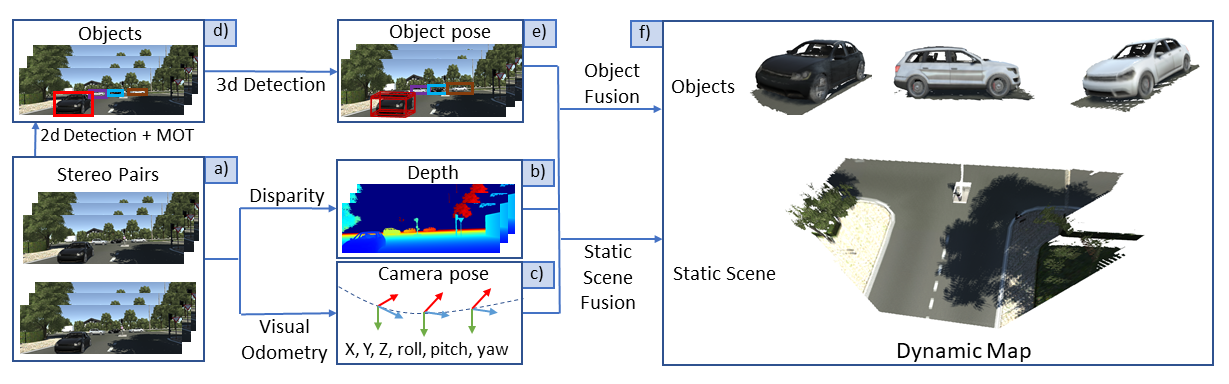}
    \vspace{-0.55cm}
    \caption{Overview of our system: (a) given stereo image pairs, we (d) estimate and track 2D bounding boxes which are used to (e) estimate pose and dimensions of 3D bounding boxes. In parallel, we estimate (b) dense depth and (c) camera poses. Finally, (f) we decompose the dense depth maps using estimated 3D bounding boxes and fuse the resulting RGB-D data into respective volumes.}
    \label{fig:overview}
    \vspace{-0.35cm}
\end{figure*}

Recent approaches have stopped treating dynamic parts of a scene as noise and started considering them as targets for 3D reconstruction or at least 2D tracking to preserve all present information. %
For instance, Dou \etal~\cite{Dou20153DSD} proposed method for dense non-rigid reconstruction, however, it was too slow for any \emph{real-time} application. 
This has been addressed by similar systems~\cite{dynamic_fusion, fusion4d}, however, such approaches are typically limited to very \emph{small-scale indoor} environments.
Kundu \etal~\cite{MulVSLAM_Abhijit_ICCV2011} proposed to reconstruct the static part of the scene and track all dynamically moving objects, however, bounding-box object tracking provide only partial information about these objects; instead we would like to maintain \emph{full 3D information} of the object over time. 

The key step for dynamic multi-body 3D reconstruction is deciding which parts of the scene should not be fused into the static volume and estimation of their 6-DoF poses.
Approaches relying solely on motion segmentation~\cite{reddy2015dynamic} are prone to ``delayed'' decisions, thus they often fuse objects that are not moving at the moment into the static volume, which requires special care when they eventually move. 
Vineet \etal~\cite{VineetMLNGPKMIP_icra15} used semantic segmentation to avoid fusing all scene parts belonging to movable class (\eg cars, cyclists, \ldots) into the static volume and simply visualised per-frame depth in the current camera view instead, hence they only suppressed the \emph{ghosting artifact} but did not reconstruct full 3D models of dynamic objects.
Recently, Barsan \etal~\cite{BarsanLPG_icra18} used stereo scene flow.
However, such methods work reliably only in the very near proximity of the car since the depth error grows quadratically with the distance~\cite{conf/ivs/LenzZGR11} which represents major limitation for self-driving scenarios where we need to perceive objects at tens or even hundreds of meters. 
Replacing stereo scene flow by its lidar-based counterpart is not straightforward due to the missing appearance information and complicated data association caused by non-uniformity and sparsity of the lidar measurements~\cite{behl2018pointflownet}. 
Moreover, the state-of-the-art methods typically take several minutes or even hours to predict.

In this paper, we propose an end-to-end system for live dense reconstruction of large-scale outdoor dynamic environments, which recovers full dense 3D information for each object. 
We represent the dynamic environment using a time-dependent map. 
We decompose the scene into the background volume corresponding to static parts of the environment and a set of independently moving objects (can be seen as analogy to ``things and stuff''~\cite{things_and_stuff}).
Each object is explicitly reconstructed in its separate and independent volume, for which we maintain relative 6-DoF poses with respect to the camera at each time-step.
Our system operates in an \emph{incremental} manner (no batch processing), handles \emph{large-scale outdoor} environments with objects at large distances, and uses \emph{computationally efficient} subsystems running in (near) real-time.
Thus, it produces the outputs on-the-fly, as the robot is moving throughout the environment which is essential to enable real-time decision making.

To address the key challenges, \ie decomposing the scene into background and independent objects, estimating their 6-DoF poses and dense depth prediction, we leverage recent advancements in data-driven approaches (supervised CNNs). 
More specifically, given a stereo video (\figref{fig:overview} A), we use a 2D object detector (\figref{fig:overview} D) to generate bounding boxes for objects present in the current frame~\cite{faster_rcnn}.  
Then we use sparse lidar measurements and Frustum PointNet~\cite{QiLWSG_cvpr18} (\figref{fig:overview} E) to estimate 3D bounding box, its dimensions and relative 6-DoF pose with respect to the current camera frame in each proposal.
The 2D proposals are fed in parallel into multi-object tracker~\cite{Lenz2015ICCV} to produce long-term tracklets for data-association.
Since we use stereo camera, we predict dense depth using PSMNet~\cite{chang2018pyramid} (\figref{fig:overview} B) and estimate camera poses using visual odometry~\cite{Geiger2011IV} (\figref{fig:overview} C).
Finally, the dense depth maps are segmented within each 3D box, and resulting ``masked'' depth maps are fused into corresponding (background or object) volumes.
We use memory efficient sparse data structures to enable dense 3D reconstruction of large-scale dynamic environments (\figref{fig:overview} F).
Note, that our data-driven approach does not require establishing multi-frame dense per-pixel correspondences, and is able to predict 3D boxes independently in each frame. 
Our approach also does not make any differences between objects that are currently moving or standing still -- this has twofold advantage: i) static map is not cluttered by spurious objects, ii) we do not need to introduce any additional mechanism converting objects from static volume into an independent one, when the previously stationary object moves. 

%% file: text/related_work.tex
\section{Related work}
\vspace{-2mm}

There is a large body of literature on incremental large-scale outdoor 3D reconstruction. 
We will focus on approaches addressing dense 3D reconstruction using stereo or lidar data for outdoor scenes with rigidly moving objects. 

\textbf{Live Static Scene Reconstruction.}
KinectFusion~\cite{NewcombeIHMKDKSHF_ismar11} represents an early approach to real-time fusion of depth data from Kinect over time to recover accurate high-quality surfaces, however, it used regular voxel grid and hence was limited to small-scale environments. 
Many methods have utilized the fact that large parts of 3D environments are empty to resolve these scalability and memory inefficiency issues. 
Some notable approaches are voxel block hashing~\cite{NiessnerZIS_tog13}, voxel hierarchy~\cite{chen2013tog} or elastic reconstruction \cite{qian2013tog} for indoor environments, and incremental dense stereo reconstruction system of Vineet \etal~\cite{VineetMLNGPKMIP_icra15} for outdoor scenes.
Recently, McCormac \etal~\cite{fusion++} proposed object-level SLAM, however, they assume static environments.

\textbf{Dynamic Scene Reconstruction.}
Vineet \etal~\cite{VineetMLNGPKMIP_icra15} proposed hash-based approach for reconstruction of large-scale dynamic environments from stereo camera.
They used semantic segmentation to avoid fusing dynamic objects in the static map and hence reduced the \emph{ghost artifact} that generally appears when fusing dynamic content into the stationary volume.
In contrast to our work, their approach is essentially not reconstructing dynamic objects in separate volumes, rather only visualises their per-frame depth in the current camera view.
Therefore all information from previous frames is lost and it is impossible to recover full spatial extend of moving objects or reason about their actions.

Similarly, Reddy \etal~\cite{ReddySCK_iros15} first perform motion segmentation to separate static and moving objects and then enforce semantic constraints in bundle adjustment for individual object reconstruction. 
This system demonstrated promising results, however, it is not able to reach (near) real-time rates and cannot scale to large environments due to expensive CRF-based optimization performed during motion segmentation and bundle adjustment.
Jian \etal~\cite{JiangPFFD_ral16} use sparse subspace clustering to segment dynamically moving objects, however this approach is also slow to be deployed in any real-world application.
Finally, co-fusion~\cite{RunzA_icra17} and MID-fusion~\cite{mid_fusion} utilize motion and semantic information to track and reconstruct static and dynamic objects from a sequence of RGBD data. 
However, most of their experiments were conducted in indoor environments using Kinect data; it needs to be shown if such methods handle large outdoor environments where distances between objects are often in tens of meters and noise present in stereo based system.

Kochanov \etal~\cite{KochanovOSL_iros16} uses stereo-based semantic segmentation and scene flow to propagate dynamic content of the scene in the map, however, such objects are not reconstructed.
Moreover, this system relies on scene flow approach of Vogel \etal~\cite{VogelSR_iccv13} which takes around 300 seconds per frame, thus making the whole system unsuitable for any real-time setup. 
Barsan \etal~\cite{BarsanLPG_icra18} recently presented a similar approach, relying on sparse scene flow and RANSAC to estimate rigid body motion of each dynamically moving object.
However, at the core of this method lies (sparse) feature matching which could pose problems for highly reflective and specular objects such as cars. 
Further, stereo scene flow methods are typically unable to estimate object pose accurately due to lack of parallax motion at distance.
They are also more susceptible to noise in depth and most state-of-the-art approaches are too slow.

\textbf{Object Pose Estimation.}
A popular approach for estimating 6-DoF object poses involves first performing motion segmentation, followed by sparse feature matching and robust RANSAC-based pose estimation~\cite{JiangPFFD_ral16} 
In this work, we instead follow a learning based approach. 
Random forest-based approaches typically required hypothesis sampling and/or were shown to work only with Kinect-like depth maps in very near distances (indoor environments)~\cite{DBLP:conf/eccv/BrachmannKMGSR14, DBLP:conf/iccv/KrullBMYGR15, DBLP:conf/cvpr/BrachmannMKYGR16}. 
In recent years, several methods based on convolution neural network (CNN) have been proposed. 
These methods typically directly generate oriented 3D boxes from single RGB image \cite{MousavianCVPR2017, Chen2015NIPS}, using RGB and point-cloud \cite{ku2018joint} or just based on point cloud \cite{QiLWSG_cvpr18}. 
In this work, we use point-cloud based Frustum PointNet model \cite{QiLWSG_cvpr18} due to its efficiency and accuracy on 3D object detection task.

%% file: text/method.tex
\vspace{-2mm}
\section{Live Reconstruction of Dynamic Worlds}
\vspace{-2mm}

Our system uses a combination of a stereo camera and lidar sensors.
The stereo cameras are intrinsically calibrated and stereo rectified to simplify disparity evaluation.
We also assume that extrinsic calibration between the two cameras and lidar is known.
Hence, we first can project the sparse lidar measurements into the camera coordinate system in each frame.
In order to decompose the scene into static background and independent (dynamic) objects, we run 2D object detector which produces ``proposals'' for estimation of 3D bounding boxes, their dimensions and 6-DoF poses from sparse lidar measurements.
The 2D proposals are also used to establish long-term data association through multi-object tracker, which runs in parallel to 3D detector, as well as dense depth and camera pose estimation.
Finally, we ``segment'' the depth maps and fuse them into their respective volumes.
The following subsections describe these parts of our reconstruction system in more detail.

\vspace{-1.5mm}
\subsection{Dynamic map representation}
\label{sec:map_representation}
\vspace{-1mm}
We represent the world using a dynamic map.
The state of our map is at each time-step $t$ encoded using the tuple of
$\calW_t = \{\bfP_t^{cw}, \bfV_t^{bg}, \calP_t^{co}, \calV_t^{o}\}$.
Here, $\bfP_t^{cw}$ is a 6-DoF rigid body pose of the camera $\bfP_t^{cw} \in {\rm \bf SE}(3)$ at \mbox{time $t$}, composed of the rotation matrix $\bfR_t^{cw} \in {\rm \bf SO}(3)$ and translation vector $\bft_t^{cw} \in \bbR^{3}$ expressed in the global reference frame.
Similarly, $\bfV_t^{bg}$ is the state of volumetric representation corresponding to background (static) part of the environment at time $t$. 
In other words, this part of the map corresponds to standard kinect fusion-like approaches.

The dynamic content of the scene is encoded using sets of object poses $\calP_t^{co}$ and corresponding relative \mbox{volumes $\calV_t^{o}$}.
Each of $N$ independent (dynamic) volumes is assigned a unique ID by the multi-object tracker.
Then, set $\calV_t^{o} = \{\bfV_{t,0}^{o}, \bfV_{t,1}^{o}, \ldots, \bfV_{t,N}^{o}\}$ encodes states of volumetric representation corresponding each of $N$ independent objects at time $t$.
Similarly, set $\calP_t^{co} = \{\bfP_{t,0}^{o}, \bfP_{t,1}^{o}, \ldots, \bfP_{t,N}^{o}\}$ represents associated relative object poses (\figref{fig:overview} F).
Note, that we describe the objects using unique IDs assigned to all $N$ objects to avoid clutter in notation, however at each time-step $t$ only objects visible in the current view frustum are present in the dynamic map.
Fig. \ref{fig:coordinate_system} illustrates different coordinate systems and their relationships for transforming points between them. 
Note, that the global reference frame can be attached to camera pose of the first frame.

\vspace{-1.5mm}
\subsection{Depth estimation}
\vspace{-1mm}
The task of dense depth estimation involves disparity prediction from calibrated and rectified stereo image pairs.
Then, we can convert disparity to depth as $d_i = \frac{b f}{z_i}$, where $d_i$ is depth at pixel $i$ corresponding to disparity value $z_i$, baseline $b$ and camera focal length $f$. 
Standard disparity estimation methods typically use some form of a prior over per-pixel matching along the scan-line and CRF-based optimization, however, more recent data-driven CNN-based approaches started providing more accurate results at faster run-times.
Thus we use in this work pyramid stereo matching network (PSM Net)~\cite{chang2018pyramid}, which represents one of the most accurate and efficient methods for disparity estimation on the KITTI benchmark.
Since data-driven methods trained on the KITTI dataset typically predict non-zero (non-invalid) values also for regions corresponding to sky (models are not penalized for such predictions during the training), we form a convex hull of lidar measurements and explicitly invalidate all depth data outside the hull.

\vspace{-1.5mm}
\subsection{Camera pose estimation}
\label{subsec_cam_pose}
\vspace{-1mm}
Given a sequence of stereo image pairs, we estimate 6-DoF camera pose $\bfP_t^{cw}$ describing rotation and translation of the camera with respect to the global reference frame at each time $t$. 
Precise camera pose estimation is important for high quality static scene reconstruction $\bfV_t^{bg}$. 
In this paper, we use LIBVISO2 library~\cite{Geiger2011IV}, which minimizes reprojection error of sparse feature matches.
In particular, this approach tiles the images into buckets to ensure the detected features are uniformly distributed across the images.
At the same time, this procedure significantly speeds-up and robustifies the matching process.
Then, the detected features are matched along epipolar lines with circular consistency check, and sporadic outliers are removed by Delaunay triangulation.
The final pose is obtained by minimising reprojection error using Gauss-Newton optimiser wrapped into RANSAC to increase robustness against outliers.
This procedure is combined with standard constant velocity Kalman filter.
Note, that our pipeline is general and any other camera pose estimation method can be used instead.

\begin{figure}[t]
    \centering
    \includegraphics[width=\linewidth]{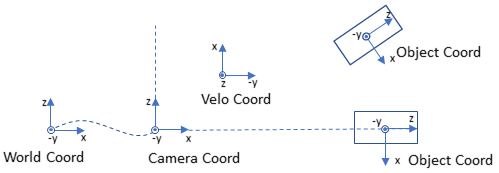}
    \caption{Coordinate systems: the camera frame is attached to the world in the first frame and data from lidar as well as recognized objects are represented using the current camera frame.}
    \label{fig:coordinate_system}
    \vspace{-0.5cm}
\end{figure}

\begin{figure*}[t]
    \centering
    \includegraphics[width=0.3295\linewidth]{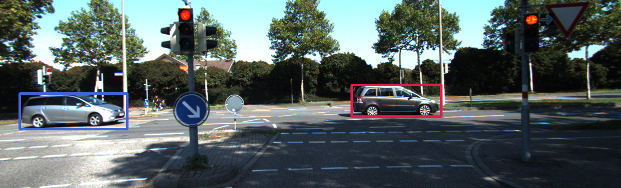}
    \includegraphics[width=0.3295\linewidth]{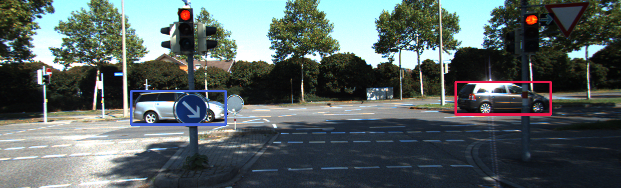}  
    \includegraphics[width=0.3295\linewidth]{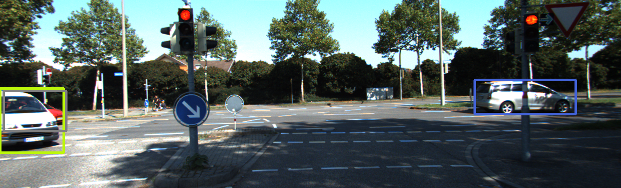} \\ \vspace{0.1cm}
    \includegraphics[width=0.3295\linewidth]{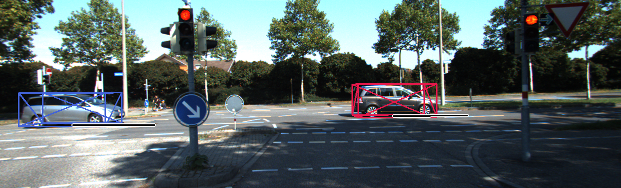}  
    \includegraphics[width=0.3295\linewidth]{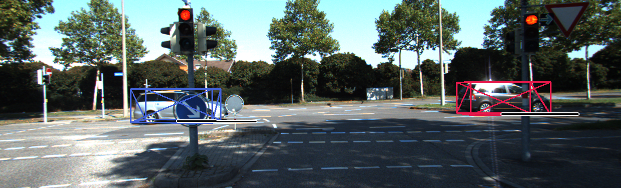} 
    \includegraphics[width=0.3295\linewidth]{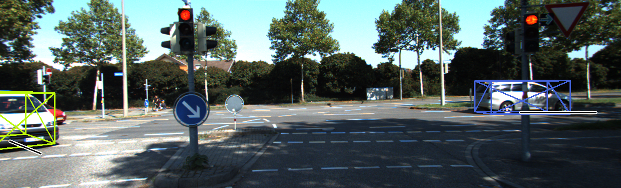}
    \caption{Examples of (dynamic) object detection, data association and 6-DoF object pose estimation. Top: 2D object detection, bottom: corresponding 3D bounding boxes. Note, that colours encode tracklet IDs (data association).}
    \label{fig:dynamic_object_detection}
    \vspace{-4mm}
\end{figure*}

\vspace{-1.5mm}
\subsection{2D object detection}
\label{subsec_2d_det}
\vspace{-1mm}

We pose estimation of independent objects present in frame $t$ as 2D object detection problem, where the goal is to describe objects of interest such as cars using tight axis-aligned 2D bounding boxes. 
Several CNN-based methods such as Faster RCNN model~\cite{faster_rcnn}, MobileNet~\cite{mobilenets} or YOLO~\cite{yolo} have been proposed. 
We use popular Faster RCNN model~\cite{faster_rcnn}, which at a high level consists of three parts: feature network, region proposal network and object detection network. 
First, the feature network, which we implement using ResNet-101~\cite{resnet}, applies a series of convolutions and ReLU non-linearities to generate a high level feature representation of input image. 
Given this feature representation, a class agnostic region proposal network (RPN) generates a set of proposals that could potentially contain objects of interest.
This typically reduces the search space to only 300 regions. 
In the final stage, the object classes, confidence scores and bounding box locations are predicted for each of these 300 boxes and non-maxima suppression is applied to generate the final set of independent objects.

\vspace{-1.5mm}
\subsection{Object pose estimation}
\label{subsec_obj_pose}
\vspace{-1mm}

The task of volumetric reconstruction of each dynamic object requires estimating accurate pose of each object. 
Let $\bfP_{t,i}^{o} \in {\rm \bf SE}(3)$ be the pose of $i^{th}$ object with respect to camera at time $t$. 
Here, pose $\bfP_{t,i}^{o}$ is composed of the rotation matrix $\bfR_{t,i}^{o} \in {\rm \bf SO}(3)$ and translation vector $\bft_{t,i}^{o} \in \bbR^{3}$ expressed in the camera reference frame (shown in Fig. \ref{fig:coordinate_system}). 

We formulate the problem of 6-DoF object pose estimation as 3D object detection problem that encloses the object of interest inside tight 3D bounding box. 
These 3D bounding boxes then provide both rotation and translation vectors with respect to the current camera frame. 
In this work, we use popular CNN-based Frustum Point Net (FPointNet) method~\cite{QiLWSG_cvpr18} which estimates 3D bounding boxes from lidar measurements in each input 2D bounding box. 

At high level, FPointNet model consists of frustum proposal, 3D instance segmentation and 3D object detection modules. 
Frustum proposal module carves out a frustum in 3D where the object of interest could be present. 
The points lying inside the frustum are segmented into objects or background using 3D instance segmentation module. 
Then, 3D box estimation module fits a tight 3D bounding box around the segmented points corresponding to the object of interest. 
Finally, object pose information $\bfP_{t,i}^{o}$ for $i^{th}$ object at time $t$ is recovered from these 3D bounding boxes.

\vspace{-1mm}
\subsection{Object tracking}
\label{sec_tracking}
\vspace{-1mm}

We use multi-object tracking-by-detection paradigm to associate object instances across time with unique IDs.
At the core of this approach lies associating detected 2D boxes in the current frame $t$ to the existing tracks. 
This is typically formulated as a labeling problem in maximum-a-posteriori (MAP) estimation framework. 
Though the problem is NP-hard, an approximate solution to the MAP problem can be found by min-cut/max-flow, however, this approach is too slow for any online setting. 
Our pipeline is based on the method of Lenz \etal~\cite{Lenz2015ICCV} that provides an efficient solution to this MAP problem, suitable for online / streaming applications. 
In particular, this method restricts the number of past frames for data association and develops an online successive shortest-path algorithm that handles streaming data without significantly affecting tracking accuracy.

\vspace{-1mm}
\subsection{Large-Scale Dynamic Scene Reconstruction}
\vspace{-1mm}
Our dynamic scene reconstruction first decomposes the scene at time $t$ into static background and $N_t$ objects (here, $N_t$ denotes a subset of objects that are present in frame $t$). 
Then, these individual RGB-D slices are integrated into the respective volumes of our dynamic map.

\textbf{Scene decomposition.}
We take a dense RGB-D map as an input, and our goal is to split it into $(N_t + 1)$ RGB-D slices ($N_t$ detected objects at time $t$ and static background).
To this end, depth data $D_t$ is first ``backprojected'' into the actual 3D points. 
Then, points lying inside space carved out by the $i$-th 3D bounding box are considered to be part of the $i$-th object. 
This is typically sufficient to decompose the scene into $N_t$ objects, however, this procedure could be enhanced \eg using segmentation output from FPointNet. 
Note that the points that do not belong to any of $N_t$ objects are automatically assigned to the static background.
It should be noted that the 3D boxes are not always accurate (tight). 
While presence of such noise may not degrade reconstruction quality of $N_t$ objects (we do not need to integrate depth map corresponding to full object at each time $t$), inclusion of points belonging to the (moving) objects into the static volume would lead to ``ghosting aritifacts''.

To avoid such issues, we approach it in a conservative way and simply invalidate the whole 2D bounding box when ``masking out'' objects for the static depth map to ensure no object points would get accidentally fused into the static volume.
Alternatively, one could rather invalidate 3D bounding box with synthetically enlarged dimensions \eg by $~15\%$. 
It is true that such strategies could introduce more holes into the reconstructed scene, however, this does not seem to represent a major problem in practice since these holes get automatically filled-in from other camera and/or object views available across the sequence.

\textbf{Hash-based TSDF fusion.}
At the core of our dynamic reconstruction lies hash-based elastic fusion, which was originally developed for static scene reconstruction~\cite{qian2013tog, qian_yi2013iccv}. 
It provides an efficient and scalable approach for integration of RGB-D data to the volumentric 3D scene representation. 
In order to handle large-scale environment, elastic fusion allocates space only for voxels that are within small distance from the perceived surface, which is measured by truncated signed distance function (TSDF)~\cite{Curless:1996:VMB:237170.237269}. Each voxel stores color and TSDF measurements. These values are updated over time by taking running average over the newly arrived color and TSDF measurements. 

\begin{figure*}[t]
    \centering
    \includegraphics[width=0.3295\linewidth]{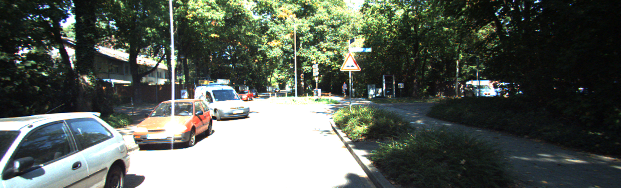}\hfill{}
    \includegraphics[width=0.3295\linewidth]{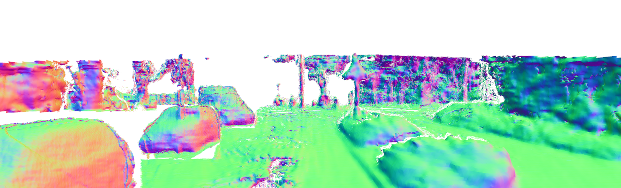}\hfill{}
    \includegraphics[width=0.3295\linewidth]{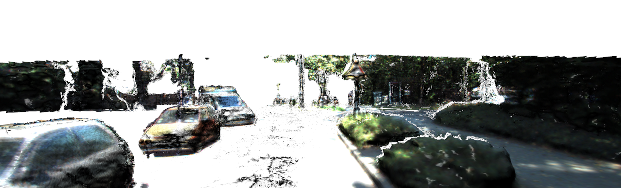} \\ \vspace{0.1cm}
    \includegraphics[width=0.3295\linewidth]{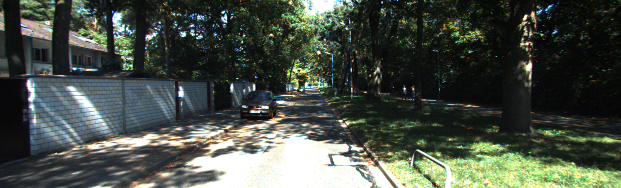}\hfill{}
    \includegraphics[width=0.3295\linewidth]{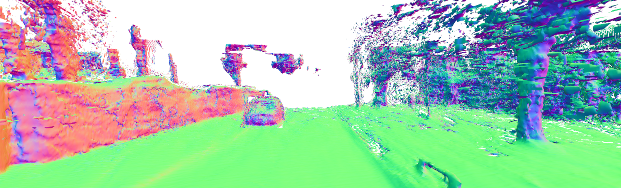}\hfill{}
    \includegraphics[width=0.3295\linewidth]{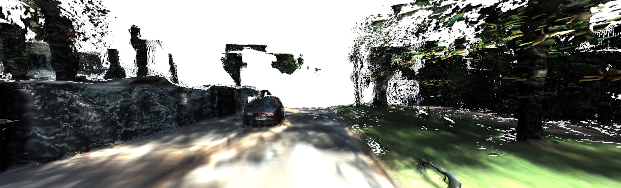} \\\vspace{0.1cm}
    \includegraphics[width=0.3295\linewidth]{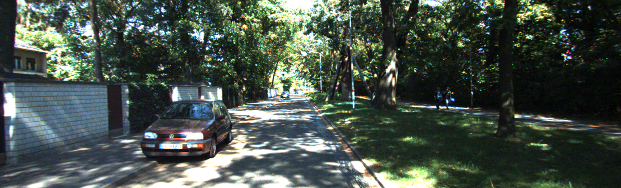}\hfill{}
    \includegraphics[width=0.3295\linewidth]{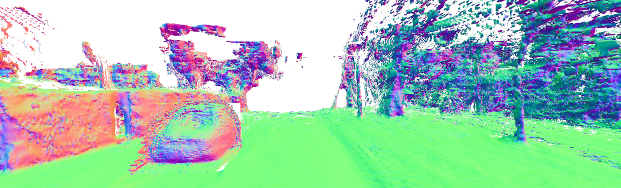}\hfill{}
    \includegraphics[width=0.3295\linewidth]{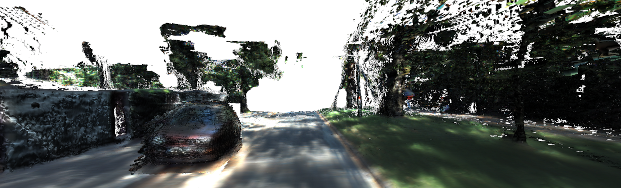} 
    \caption{Typical output of our system: (left) input images, (middle) Phong-shaded surface normals, (right) Phong-shaded textured model.}
    \label{fig:normals}
    \vspace{-4mm}
\end{figure*}

\textbf{Management of a dynamic map.}
As we have mentioned in $\S$\ref{sec:map_representation}, our dynamic map $\calW_t$ is represented by two components. First, it incorporates information about static volume $\bfV_t^{bg}$, and camera poses $\bfP_t^{cw}$ at each time $t$ forming a camera trajectory. 
Second, it also consists of a set of $N_t$ independent volumes $\calV_t^{o}$ corresponding to objects and their associated relative poses $\calP_t^{co}$ in camera frame reference.

Reconstruction process starts by allocating the static volume. At each time $t$, the masked depth map for the static scene is integrated into the static volume $\bfV_t^{bg}$ using camera pose $\bfP_t^{cw}$ (estimated in $\S$\ref{subsec_cam_pose}). The dynamic map $\calW_t$ is updated with the static volume and camera pose at time $t$. 

To reconstruct each object present at time $t$, we begin by checking if the object volume exists in the dynamic map $\calW_t$. 
If such a volume is not found, we first allocate a new volume $\bfV_t^{o}$ for this object. 
Then, we integrate a segmented RGB-D slice for the object into the volume $\bfV_t^{o}$ using object pose $\bfP_t^{co}$ (recovered in $\S$\ref{subsec_obj_pose}). 
Note that this pose $\bfP_t^{co}$ is always relative to the current camera frame. 
Thus the objects are reconstructed directly without need to explicitly compute position and rotation in the global coordinate frame. Hence potential drift of camera pose estimation does not influence quality of dense 3D reconstruction of dynamic objects (of course, the absolute pose of an object in the global coordinate frame is still dependent on camera trajectory).

\textbf{Visualisation.}
\label{sec:dynamic_map_visualization}
At each time $t$, the current state of the dynamic map $\calW_t$ is being visualised by first placing dynamic objects into their respective locations within the world coordinate system. This involves expressing points in the object frame to the world frame by applying projection using  $\bfP_t^{ow}=\bfP_t^{cw} \bfP_t^{o}$. 
The current state of dynamic map $\calW_t$ then can be visualised from arbitrary camera, we typically use camera-pose $\bfP_t^{cw}$ at time $t$ to get the actual live view. 

We could also visualise not just the ``on-the-fly'' live view, as seen from the moving camera, but also the virtual (off-line) ``fly through the reconstructed scene'' (\ie the final state of volumes). 
This can be done simply by using the final state of all volumes in all frames of the sequence.

%% file: text/results.tex
\vspace{-3mm}
\section{Results}
\vspace{-2mm}
We use KITTI dataset \cite{Geiger2012CVPR} to evaluate our approach.  
The KITTI dataset contains a variety of challenging outdoor sequences containing many moving objects such as cars. 
These sequences were captured in residential, city and highway areas. 
Imaging sensors include two colour and two grayscale PointGrey Flea2 cameras and Velodyne \mbox{HDL-64E} lidar.
All sequences were captured at a resolution of $1241\times376$ pixels using both camera and lidar.
Both sensors scan the environment at 10 Hz and cameras trigger when the spinning lidar is oriented in the same direction as cameras. 
We use stereo inputs from colour stereo cameras and lidar data for all experiments. 
Cameras were intrinsically and extrinsically calibrated, all images stereo rectified and lidar to cameras mapping is known.

We demonstrate both the qualitative and quantitative results on four diverse sequences from the KITTI tracking data. 
For quantitative evaluation, we use per-frame sparse lidar measurements as ground-truth data (note that is not perfect as we cannot ``untwist'' the measurements of spinning lidar for moving objects, but it is the best available real-world data for evaluation). 
We report the standard mean relative error (MRE) metric, which is defined as $\frac{1}{M}\sum_i^M |d_i-d_i^{gt}| / d_i^{gt}$.
Here, $d_i$ and $d_i^{gt}$ are respectively reconstructed and ground truth depth normalized over $M$ valid lidar points.
The MRE error measures relative per-pixel error, \ie an error of $0.1$m at depth of $1$m is penalized equally to an error of $1$m at a depth of $10$m. 
We compare our approach with a standard fusion based reconstruction method that does not explicitly handle multi-body dynamic objects.  
To make the comparison fair, we have adapted the baseline to large-scale outdoor environments, \ie it uses the very same visual odometry, dense depth and parameters as our approach.
We have used voxels of $4.68$cm for the background volume and $1.56$cm for objects.
The depth considered for reconstruction was truncated at $40$m.
Our system is implemented on top of the Open3D library~\cite{Zhou2018} and we \textbf{released the full source code at}~\mbox{\url{https://github.com/omiksik/dfusion}} 

\begin{figure*}[t]
    \centering
    \includegraphics[width=\linewidth]{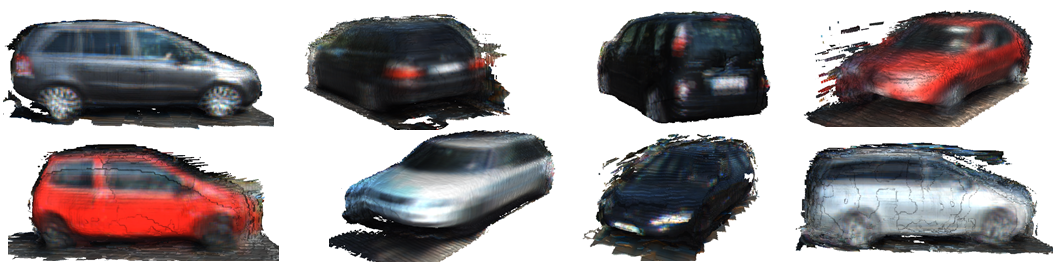}
    \caption{Examples of reconstructed objects.}
    \label{fig:reconstructed_objects}
\end{figure*}

\begin{figure*}[t]
    \centering
    \includegraphics[width=\linewidth]{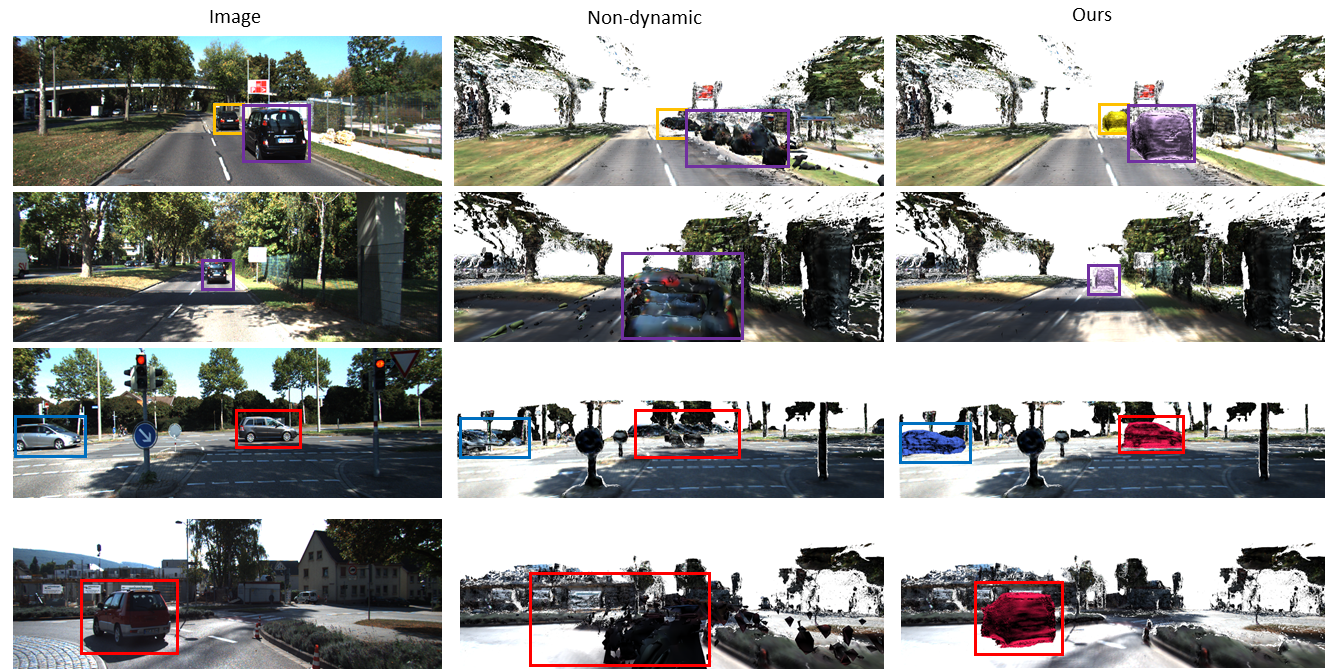} \\
    \includegraphics[width=\linewidth]{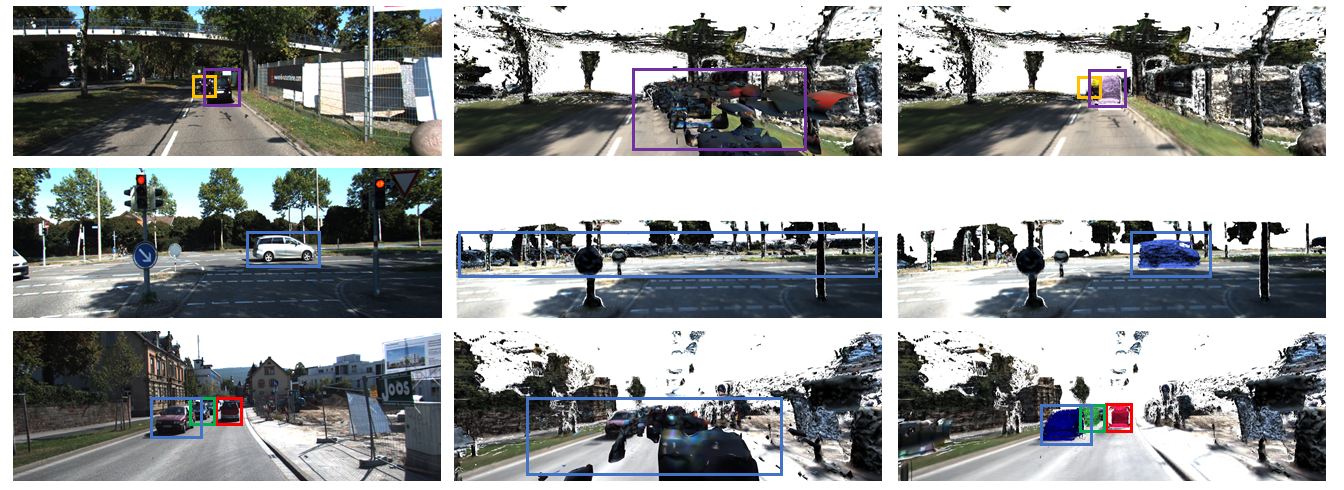}
    \caption{Typical outputs of our system: (left) input images with detected bounding boxes associated over time (trajectories coded by different colours), (middle) Phong-shaded textured output, (right) in our approach, each object is explicitly treated as an independent instance reconstructed in a separate volume; here we visualise it by using colour-coding corresponding to a particular tracklet ID.}
    \label{fig:examples}
\end{figure*}

\begin{figure*}[t]
    \centering
    \includegraphics[width=0.3295\linewidth]{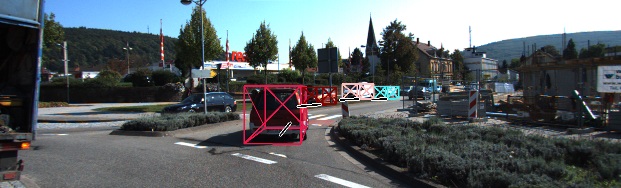}\hfill{}
    \includegraphics[width=0.3295\linewidth]{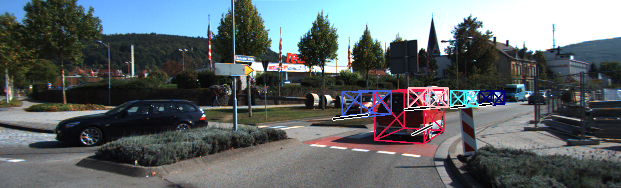}\hfill{}
    \includegraphics[width=0.3295\linewidth]{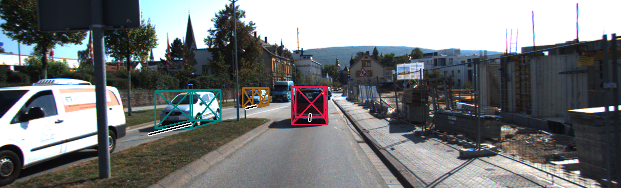}\\\vspace{0.1cm}
    \includegraphics[width=0.3295\linewidth]{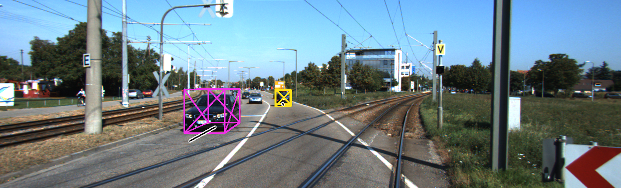}\hfill{}
    \includegraphics[width=0.3295\linewidth]{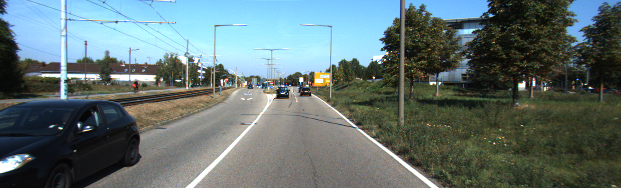}\hfill{}
    \includegraphics[width=0.3295\linewidth]{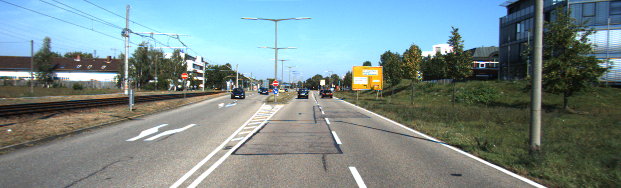}
    \caption{Failure cases: whenever the detector misses an object, this part of the image would get fused into the static volume, hence would lead to ``ghosting artifact'' for objects which are moving.}
    \label{fig:failures}
    \vspace{-3mm}
\end{figure*}

\textbf{Qualitative Evaluation.}
We qualitatively illustrate the impact of our approach in recovering high quality reconstruction of scenes with multiple moving objects. 
In all figures, we show ``on-the-fly'' output, that is the output as seen from the camera view as it is moving (\cf $\S$\ref{sec:dynamic_map_visualization}).

In \figref{fig:examples}, we show more examples of how our approach accurately reconstructs moving objects along the static scene. 
Note the objects marked inside rectangular boxes, and how they are properly reconstructed using our multi-body dynamic fusion approach (right column), compared to standard non-dynamic fusion where both the static and dynamic objects are fused in the same map, which leads to {\em ghost artifact}. 
Such {\em ghost artifact} can be clearly seen on several examples taken from all four sequences shown in the middle column of the \figref{fig:examples}.
It should be emphasized, that even if the output of non-dynamic fusion and our approach look the same when cars are not moving, the key difference between the two is that we explicitly reconstruct each instance it its own volume which i) simplifies reasoning about object actions and ii) does not require any extra mechanism when the car starts moving.
To better illustrate quality of reconstructed objects, we show multiple examples in \figref{fig:reconstructed_objects} and also the surface normals in~\figref{fig:normals}.
Finally \figref{fig:failures} illustrates most common failure cases, \ie situation when the object detector misses some cars present in the scene.
Full sequences are shown in the supplementary video at \url{https://youtu.be/gCCVwE3vI-E} 

\begin{table}[t]
    \centering
 \caption{Mean relative error computed across whole image}
   \footnotesize
 \begin{tabular}{c  c  c  c  c}
  & Scene 1 & Scene 2 & Scene 3 & Scene 4 \\ [0.5ex]
 \hline 
 \multicolumn{5}{c}{10 meters} \\ \hline
Non-Dynamic & 0.20 & 0.19  & 0.35 & 0.45 \\ 
 Dynamic & 0.18 & \textbf{0.15}  & 0.35 & \textbf{0.36} \\
 \hline 
 \multicolumn{5}{c}{20 meters} \\ \hline
 Non-Dynamic & 0.25  & 0.25  & 0.26  & 0.30  \\ 
 Dynamic & 0.24 & \textbf{0.20}  & 0.28 & 0.32 \\
 \hline 
 \multicolumn{5}{c}{30 meters} \\ \hline
 Non-Dynamic & 0.26 & 0.25  & 0.26 & 0.39  \\
 Dynamic & 0.25 & \textbf{0.20}  & 0.26 & \textbf{0.30}\\
 \hline 
 \multicolumn{5}{c}{40 meters} \\ \hline
 Non-Dynamic & 0.26 & 0.25  & 0.26 & 0.37 \\ 
 Dynamic & 0.25 & \textbf{0.20}  & 0.26 & \textbf{0.30}  \\
 \hline
\end{tabular}
\label{tab:dynamic_depth_overall}
\vspace{-5mm}
\end{table}

\textbf{Quantitative Evaluation.}
Here we provide the quantitative evaluation for non-dynamic and proposed dynamic approaches.
We consider pixels with ground-truth depth within $10$m, $20$m, $30$m and $40$m for quantitative evaluation.
In Tab. \ref{tab:dynamic_depth_overall}, we report the standard mean relative error (MRE) evaluated across the whole image.
In order to better highlight the quality of reconstructed objects, we also evaluate our method on points lying inside the 2D object bounding boxes as shown in Tab.~\ref{tab:dynamic_depth_object}. 
On most of the sequences, we can see that our dynamic reconstruction approach achieves a significant improvement of almost $2\times$ reduced error on estimated depth for dynamic objects compared to the baseline method with non-dynamic fusion. 
Overall, the presented approach for multi-body reconstruction achieves almost $5\%$ to $25\%$ improvement in depth accuracy over the baseline method on different sequences. 
This suggests that the dynamic reconstruction is important for alleviating {\em ghost artifact} in the reconstruction.
The only sequence, on which do not achieve better results is Scene 3, which is mostly static. 
Thus it is not surprising our method achieves slightly worse results, however, we treat individual objects as independent instances in contrast to non-dynamic fusion approach.

\textbf{Timing details.}
The presented approach runs at around 2fps. 
The main limiting factor is PSMNet based stereo estimation which takes around 410ms for every stereo pair.
The other parts of the pipeline are relatively faster to evaluate; in particular visual odometry runs at 50ms, MOT tracker at 10ms, Faster-RCNN at 79ms and FPointNet at 170ms. Further, these components run in parallel to PSMNet.

\begin{table}[t]
    \centering
 \caption{Mean relative error computed within bounding boxes}
\footnotesize
\begin{tabular}{c  c  c  c  c}
  & Scene 1 & Scene 2 & Scene 3 & Scene 4 \\ [0.5ex]
 \hline
 \multicolumn{5}{c}{10 meters} \\ \hline
 Non-Dynamic & 0.92 & 0.56  & \textbf{0.38} & 0.84 \\ 
 Dynamic & \textbf{0.43}  & \textbf{0.39}  & 0.44 & \textbf{0.36} \\
 \hline 
 \multicolumn{5}{c}{20 meters} \\ \hline
 Non-Dynamic & 0.53  & 0.45  & 0.29 & 0.68 \\ 
 Dynamic & \textbf{0.35} & \textbf{0.27}  & 0.31 & \textbf{0.33} \\
 \hline 
  \multicolumn{5}{c}{30 meters} \\ \hline
 Non-Dynamic & 0.51 & 0.43  & 0.28  & 0.63  \\ 
 Dynamic & \textbf{0.34} & \textbf{0.26}  & 0.29 & \textbf{0.31} \\
 \hline
 \multicolumn{5}{c}{40 meters} \\ \hline
 Non-Dynamic & 0.51 & 0.42  & 0.27 & 0.60 \\ 
 Dynamic & \textbf{0.34} & \textbf{0.27} & 0.29 & \textbf{0.31} \\
 \hline
\end{tabular}
\label{tab:dynamic_depth_object}
\vspace{-5mm}
\end{table}

%% file: text/conclusion.tex
\vspace{-2mm}
\section{Conclusion}
\vspace{-2mm}
We have presented an end-to-end system for live reconstruction of large scale dynamic scenes. 
The key observation is that the 6-DoF object pose estimation for dynamic scene reconstruction can be framed within 3D object detection framework. 
Such framework helped to represent whole scene using a time-dependent dynamic map, in which each object is explicitly reconstructed in its own independent volume. 
We have demonstrated high quality reconstruction of static and dynamic objects on various KITTI sequences. 

The presented approach could benefit from improvements in efficiency and accuracy of depth estimation, 2D and 3D detection or multi-object tracking. 
Further, temporal smoothness in object pose estimation would help to regularize reconstruction of each individual object.

%% file: main.bbl
\begin{thebibliography}{10}\itemsep=-1pt

\bibitem{things_and_stuff}
E.~H. Adelson.
\newblock On seeing stuff: the perception of materials by humans and machines.
\newblock In {\em Proc. SPIE}, 2001.

\bibitem{rome_in_a_day}
S.~Agarwal, Y.~Furukawa, N.~Snavely, I.~Simon, B.~Curless, S.~M. Seitz, and
  R.~Szeliski.
\newblock Building rome in a day.
\newblock {\em Communications of the ACM}, 2011.

\bibitem{BarsanLPG_icra18}
I.~A. Barsan, P.~Liu, M.~Pollefeys, and A.~Geiger.
\newblock Robust dense mapping for large-scale dynamic environments.
\newblock In {\em ICRA}, 2018.

\bibitem{behl2018pointflownet}
A.~Behl, D.~Paschalidou, S.~Donné, and A.~Geiger.
\newblock Pointflownet: Learning representations for rigid motion estimation
  from point clouds.
\newblock In {\em arXiv}, 2018.

\bibitem{dynslam}
B.~Besc{\'{o}}s, J.~M. F{\'{a}}cil, J.~Civera, and J.~Neira.
\newblock Dynslam: Tracking, mapping and inpainting in dynamic scenes.
\newblock {\em CoRR}, abs/1806.05620, 2018.

\bibitem{DBLP:conf/eccv/BrachmannKMGSR14}
E.~Brachmann, A.~Krull, F.~Michel, S.~Gumhold, J.~Shotton, and C.~Rother.
\newblock Learning 6d object pose estimation using 3d object coordinates.
\newblock In {\em ECCV}, 2014.

\bibitem{DBLP:conf/cvpr/BrachmannMKYGR16}
E.~Brachmann, F.~Michel, A.~Krull, M.~Y. Yang, S.~Gumhold, and C.~Rother.
\newblock Uncertainty-driven 6d pose estimation of objects and scenes from a
  single {RGB} image.
\newblock In {\em CVPR}, 2016.

\bibitem{chang2018pyramid}
J.-R. Chang and Y.-S. Chen.
\newblock Pyramid stereo matching network.
\newblock In {\em CVPR}, 2018.

\bibitem{chen2013tog}
J.~Chen, D.~Bautembach, and S.~Izadi.
\newblock Scalable real-time volumetric surface reconstruction.
\newblock {\em ACM-TOG}, 2013.

\bibitem{Chen2015NIPS}
X.~Chen, K.~Kundu, Y.~Zhu, A.~Berneshawi, H.~Ma, S.~Fidler, and R.~Urtasun.
\newblock 3d object proposals for accurate object class detection.
\newblock In {\em NIPS}, 2015.

\bibitem{Curless:1996:VMB:237170.237269}
B.~Curless and M.~Levoy.
\newblock A volumetric method for building complex models from range images.
\newblock In {\em SIGGRAPH}, 1996.

\bibitem{fusion4d}
M.~Dou, S.~Khamis, Y.~Degtyarev, P.~Davidson, S.~R. Fanello, A.~Kowdle, S.~O.
  Escolano, C.~Rhemann, D.~Kim, J.~Taylor, P.~Kohli, V.~Tankovich, and
  S.~Izadi.
\newblock Fusion4d: Real-time performance capture of challenging scenes.
\newblock {\em ACM TOG}, 2016.

\bibitem{Dou20153DSD}
M.~Dou, J.~Taylor, H.~Fuchs, A.~W. Fitzgibbon, and S.~Izadi.
\newblock 3d scanning deformable objects with a single rgbd sensor.
\newblock {\em CVPR}, 2015.

\bibitem{Geiger2012CVPR}
A.~Geiger, P.~Lenz, and R.~Urtasun.
\newblock Are we ready for autonomous driving? the kitti vision benchmark
  suite.
\newblock In {\em CVPR}, 2012.

\bibitem{Geiger2011IV}
A.~Geiger, J.~Ziegler, and C.~Stiller.
\newblock Stereoscan: Dense 3d reconstruction in real-time.
\newblock In {\em IVS}, 2011.

\bibitem{resnet}
K.~He, X.~Zhang, S.~Ren, and J.~Sun.
\newblock Deep residual learning for image recognition.
\newblock In {\em CVPR}, 2016.

\bibitem{mobilenets}
A.~G. Howard, M.~Zhu, B.~Chen, D.~Kalenichenko, W.~Wang, T.~Weyand,
  M.~Andreetto, and H.~Adam.
\newblock Mobilenets: Efficient convolutional neural networks for mobile vision
  applications.
\newblock {\em CoRR}, abs/1704.04861, 2017.

\bibitem{JiangPFFD_ral16}
C.~Jiang, D.~P. Paudel, Y.~D. Fougerolle, D.~Fofi, and C.~Demonceaux.
\newblock Static-map and dynamic object reconstruction in outdoor scenes using
  3-d motion segmentation.
\newblock {\em RA-Letters}, 2016.

\bibitem{KochanovOSL_iros16}
D.~Kochanov, A.~Osep, J.~St{\"{u}}ckler, and B.~Leibe.
\newblock Scene flow propagation for semantic mapping and object discovery in
  dynamic street scenes.
\newblock In {\em IROS}, 2016.

\bibitem{DBLP:conf/iccv/KrullBMYGR15}
A.~Krull, E.~Brachmann, F.~Michel, M.~Y. Yang, S.~Gumhold, and C.~Rother.
\newblock Learning analysis-by-synthesis for 6d pose estimation in {RGB-D}
  images.
\newblock In {\em ICCV}, 2015.

\bibitem{ku2018joint}
J.~Ku, M.~Mozifian, J.~Lee, A.~Harakeh, and S.~Waslander.
\newblock Joint 3d proposal generation and object detection from view
  aggregation.
\newblock {\em IROS}, 2018.

\bibitem{MulVSLAM_Abhijit_ICCV2011}
A.~Kundu, K.~M. Krishna, and C.~V. Jawahar.
\newblock Realtime multibody visual slam with a smoothly moving monocular
  camera.
\newblock In {\em CVPR}, 2011.

\bibitem{Lenz2015ICCV}
P.~Lenz, A.~Geiger, and R.~Urtasun.
\newblock Followme: Efficient online min-cost flow tracking with bounded memory
  and computation.
\newblock In {\em ICCV}, 2015.

\bibitem{conf/ivs/LenzZGR11}
P.~Lenz, J.~Ziegler, A.~Geiger, and M.~Roser.
\newblock Sparse scene flow segmentation for moving object detection in urban
  environments.
\newblock In {\em Intelligent Vehicles Symposium}, 2011.

\bibitem{fusion++}
J.~McCormac, R.~Clark, M.~Bloesch, A.~J. Davison, and S.~Leutenegger.
\newblock Fusion++: Volumetric object-level {SLAM}.
\newblock {\em CoRR}, abs/1808.08378, 2018.

\bibitem{MousavianCVPR2017}
A.~Mousavian, D.~Anguelov, J.~Flynn, and J.~Kosecka.
\newblock 3d bounding box estimation using deep learning and geometry.
\newblock In {\em CVPR}, 2017.

\bibitem{dynamic_fusion}
R.~A. Newcombe, D.~Fox, and S.~M. Seitz.
\newblock Dynamicfusion: Reconstruction and tracking of non-rigid scenes in
  real-time.
\newblock In {\em CVPR}, 2015.

\bibitem{NewcombeIHMKDKSHF_ismar11}
R.~A. Newcombe, S.~Izadi, O.~Hilliges, D.~Molyneaux, D.~Kim, A.~J. Davison,
  P.~Kohli, J.~Shotton, S.~Hodges, and A.~W. Fitzgibbon.
\newblock Kinectfusion: Real-time dense surface mapping and tracking.
\newblock In {\em ISMAR}, 2011.

\bibitem{NiessnerZIS_tog13}
M.~Nie{\ss}ner, M.~Zollh{\"{o}}fer, S.~Izadi, and M.~Stamminger.
\newblock Real-time 3d reconstruction at scale using voxel hashing.
\newblock {\em {ACM} TOG}, 2013.

\bibitem{QiLWSG_cvpr18}
C.~R. Qi, W.~Liu, C.~Wu, H.~Su, and L.~J. Guibas.
\newblock Frustum pointnets for 3d object detection from {RGB-D} data.
\newblock In {\em CVPR}, 2018.

\bibitem{reddy2015dynamic}
N.~D. Reddy, P.~Singhal, V.~Chari, and K.~M. Krishna.
\newblock Dynamic body vslam with semantic constraints.
\newblock {\em IROS}, 2015.

\bibitem{ReddySCK_iros15}
N.~D. Reddy, P.~Singhal, V.~Chari, and K.~M. Krishna.
\newblock Dynamic body {VSLAM} with semantic constraints.
\newblock In {\em IROS}, 2015.

\bibitem{yolo}
J.~Redmon and A.~Farhadi.
\newblock Yolov3: An incremental improvement.
\newblock {\em CoRR}, abs/1804.02767, 2018.

\bibitem{faster_rcnn}
S.~Ren, K.~He, R.~Girshick, and J.~Sun.
\newblock Faster r-cnn: Towards real-time object detection with region proposal
  networks.
\newblock In {\em NIPS}, 2015.

\bibitem{RunzA_icra17}
M.~R{\"{u}}nz and L.~Agapito.
\newblock Co-fusion: Real-time segmentation, tracking and fusion of multiple
  objects.
\newblock In {\em ICRA}, 2017.

\bibitem{VineetMLNGPKMIP_icra15}
V.~Vineet, O.~Miksik, M.~Lidegaard, M.~Nie{\ss}ner, S.~Golodetz, V.~A.
  Prisacariu, O.~K{\"{a}}hler, D.~W. Murray, S.~Izadi, P.~P{\'{e}}rez, and
  P.~H.~S. Torr.
\newblock Incremental dense semantic stereo fusion for large-scale semantic
  scene reconstruction.
\newblock In {\em ICRA}, 2015.

\bibitem{VogelSR_iccv13}
C.~Vogel, K.~Schindler, and S.~Roth.
\newblock Piecewise rigid scene flow.
\newblock In {\em ICCV}, 2013.

\bibitem{reconstructing_museums}
J.~Xiao and Y.~Furukawa.
\newblock Reconstructing the world's museums.
\newblock {\em IJCV}, 2014.

\bibitem{mid_fusion}
B.~Xu, W.~Li, D.~Tzoumanikas, M.~Bloesch, A.~J. Davison, and S.~Leutenegger.
\newblock Mid-fusion: Octree-based object-level multi-instance dynamic {SLAM}.
\newblock {\em CoRR}, abs/1812.07976, 2018.

\bibitem{qian2013tog}
Q.-Y. Zhou and V.~Koltun.
\newblock Dense scene reconstruction with points of interest.
\newblock {\em ACM-TOG}, 2013.

\bibitem{qian_yi2013iccv}
Q.-Y. Zhou, S.~Miller, and V.~Koltun.
\newblock Elastic fragments for dense scene reconstruction.
\newblock In {\em ICCV}, 2013.

\bibitem{Zhou2018}
Q.-Y. Zhou, J.~Park, and V.~Koltun.
\newblock {Open3D}: {A} modern library for {3D} data processing.
\newblock {\em arXiv:1801.09847}, 2018.

\end{thebibliography}
